\documentclass[11pt,a4paper]{article}
\usepackage[hyperref]{acl2021}
\usepackage{times}
\usepackage{pifont}
\usepackage{amsmath}

\usepackage{latexsym}

\usepackage{booktabs}
\usepackage{multicol}
\usepackage{multirow}
\usepackage{graphicx}
\usepackage{amsmath}
\usepackage{enumitem}
\usepackage{todonotes}
\usepackage{microtype}
\usepackage{fancyhdr}

\newcommand{\comment}[1]{}
\newcommand{\tabitem}{~~\llap{\textbullet}~~}

\usepackage{color, colortbl}
\definecolor{LightCyan}{rgb}{0.88,1,1}

\aclfinalcopy 

\newcommand*\samethanks[1][\value{footnote}]{\footnotemark[#1]}

\title{Learning to Automate Follow-up Question Generation using Process Knowledge for Depression Triage on Reddit Posts}

\author{ Shrey Gupta\textsuperscript{\rm 1}\thanks{\;Authors contributed equally}\ , Anmol Agarwal\textsuperscript{\rm 1}\samethanks\ , Manas Gaur\textsuperscript{\rm 2}, \\ \textbf{Kaushik Roy}\textsuperscript{\rm 2}, \textbf{Vignesh Narayanan}\textsuperscript{\rm 2}, \textbf{Ponnurangam Kumaraguru}\textsuperscript{\rm 1}, \textbf{Amit Sheth}\textsuperscript{\rm 2}\\
\textsuperscript{\rm 1}International Institute of Information Technology, Hyderabad, India \\
\texttt{\{shrey.gupta, anmol.agarwal\}@students.iiit.ac.in}, \\ \texttt{pk.guru@iiit.ac.in} \\
\textsuperscript{\rm 2}AI Institute, University of South Carolina, SC, USA \\
\texttt{\{mgaur, kaushikr\}@email.sc.edu}, \texttt{\{vignar, amit\}@sc.edu} \\
}

\date{}

\begin{document}
\maketitle

\begin{abstract}
Conversational Agents (CAs) powered with deep language models (DLMs) have shown tremendous promise in the domain of mental health. Prominently, the CAs have been used to provide informational or therapeutic services (e.g., cognitive behavioral therapy) to patients. However, the utility of CAs to assist in mental health triaging has not been explored in the existing work as it requires a controlled generation of follow-up questions (FQs), which are often initiated and guided by the mental health professionals (MHPs) in clinical settings. 
In the context of `depression', our experiments show that DLMs coupled with process knowledge in a mental health questionnaire generate 12.54\% and 9.37\% better FQs based on similarity and longest common subsequence matches to questions in the PHQ-9 dataset respectively, when compared with DLMs without process knowledge support.
Despite coupling with process knowledge, we find that DLMs are still prone to hallucination, i.e., generating redundant, irrelevant, and unsafe FQs.  We demonstrate the challenge of using existing datasets to train a DLM for generating FQs that adhere to clinical process knowledge. To address this limitation, we prepared an extended PHQ-9 based dataset, PRIMATE, in collaboration with MHPs. PRIMATE contains annotations regarding whether a particular question in the PHQ-9 dataset has already been answered in the user's initial description of the mental health condition. 
We used PRIMATE to train a DLM in a supervised setting to identify which of the PHQ-9 questions can be answered directly from the user's post and which ones would require more information from the user. Using performance analysis based on MCC scores, we show that PRIMATE is appropriate for identifying questions in PHQ-9 that could guide generative DLMs towards controlled FQ generation (with minimal hallucination) suitable for aiding triaging. The dataset created as a part of this research can be obtained from \href{https://github.com/primate-mh/Primate2022}{here}.
\end{abstract}

\section{Introduction}
Conversational agents (CAs) powered by DLMs are software designed to interact with human users for specific tasks. For mental health purposes, particularly depression, CAs have been studied extensively in prior work for helping patients follow generic mental health guidelines, typically by providing reminders to assist patients in adhering to the medication and therapy strategy outlined by a mental health professional (MHP)\footnote{\url{https://tinyurl.com/yfp3bhr2}}\footnote{\url{https://woebothealth.com/}}. However, previous work on depression have not examined the use of CAs for triaging. For the purpose of triaging, CAs should learn to generate controlled and clinical process knowledge-guided discourse that can assist MHPs in diagnosis. Our research suggests a clinically grounded and explainable methodology to develop conversational information-seeking tools, first to learn “what symptoms the user is suffering” and “what extra information is needed for triaging.”

\begin{figure*}[!ht]
    \centering
    \includegraphics[width=150mm, scale=0.50]{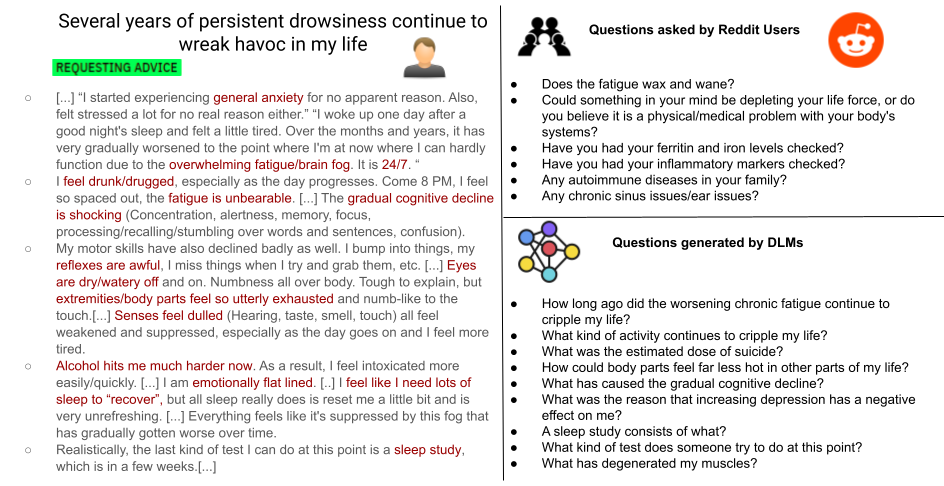}
    \caption{
    Reddit is a rich source for bringing crowd perspective in training DLMs over conversational data. On the \textbf{left} is a sample post from r/depression\_help which sees inquisitive interaction from other Reddit users. At the \textbf{top-right} are the FQs asked by the Reddit users in the comments. These FQs are aimed at understanding the severity of the mental health situation of the user and are hence, diagnostically relevant. At the \textbf{bottom-right} are the questions generated by DLMs. It can be seen that these are not suitable FQs.}
    \label{fig:motivate}
\end{figure*}

CAs are susceptible to irrelevant and sometimes harmful questions when generating FQs or responses to a patient suffering from depression \cite{miner2016smartphone}. The primary reason for irrelevant and harmful questions is that CAs cannot incorporate contextual information in generating appropriate follow-up questions (FQs) (see Figure \ref{fig:motivate}). Further, the sensitivity of the conversation and a controlled generation process are essential characteristics of patient-clinician interactions, which are difficult to embed in DLM-based CAs. Therefore, question generation (QG) in mental health is challenging, and research to develop CAs for automating triage has not been explored.

Procedures for generating semantically related and logically ordered questions in the mental health domain are a form of process knowledge manifested in various clinical instruments for mental health triage. For example, the severity of depression is measured using Patient Health Questionnaire (PHQ-9). Enforcing DLMs to follow process knowledge, like in PHQ-9, would make CAs generate FQs similar to an MHP when they are seeking information from the patient \cite{karasz2012we}. Unfortunately, datasets that meet this criterion are currently unavailable. Though clinical diagnostic interviews exist, they are not rich, sufficiently dense, and varied to train DLMs \cite{manas2021knowledge,gratch2014distress}. Further, we require dataset(s) that includes \textit{support seeking queries} and \textit{natural questions} that show help providing behavior. For this purpose, anonymized user-generated conversational data in Mental Health support communities on Reddit provides a rich source of fine-grained, contextual, and diverse information suitable for fine-tuning DLMs. Specific to depression, we explored posts and comments in r/depression\_help.

\begin{figure*}[!htbp]
    \centering
    \includegraphics[width=120mm, scale=0.5]{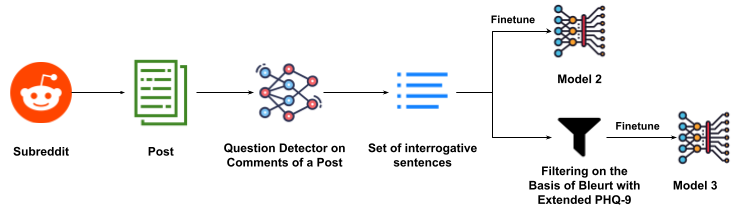}
    \caption{An illustration of our pipeline for developing Model 2 and Model 3 using T5 as the deep language models. Starting with posts (including comments) from r/depression\_help, we filter out comments that are neither interrogative nor information seeking in nature to yield a posts-questions dataset for fine-tuning T5. This dataset was further filtered using extended PHQ-9 before using it to fine-tune T5 (Model 3).}
    \label{fig:architecture}
\end{figure*}

In the current research, we emphasize the limitations of T5, a state-of-the-art DLM\footnote{Current DLMs are either variants of T5 or built from T5} to generate process knowledge-like FQs using the data from r/depression\_help \cite{raffel2019exploring}. We filtered the dataset by retaining only posts with at least one comment that seeks additional information from the user seeking support. Further filtering of comments was performed using PHQ-9 to assist T5 in generating relevant FQs (see Figure \ref{fig:architecture}). We found that the outcome is substantial for the single turn question answering model; however, not suitable for mental health triage, which is a discourse. We conducted a series of experiments keeping our focus on 'depression' and leveraged its associated process knowledge for mental health triage: the PHQ-9 \cite{kroenke2001phq}. 
To the best of our knowledge, FQ generation relating to \textit{depression} has never been studied using PHQ-9 for \textit{discourse modeling} and \textit{generation}.

We make the following key contributions: 
(a) \textbf{Extending PHQ-9:} PHQ-9 questions are limited in scope for common NLP tasks like finetuning. In collaboration with MHPs, we prepared a list of 134 sub-questions for nine PHQ-9 questions for better fine-tuning of T5.
(b) We analyzed the performance of three variants of T5 using BLEURT \cite{sellam2020bleurt} and ROUGE-L scores that measure semantic relatedness and exact match similarity of generated question to sub-questions of PHQ-9. 
(c) \textbf{PRIMATE Dataset:} Lessons learned during our experiments suggested that T5 must be trained in a supervised setting to capture `what the user has already mentioned about his/her depression condition in the post-text' and then generate FQs. Along with MHPs, we constructed a novel \textbf{PRIMATE} (\textbf{PR}ocess knowledge \textbf{I}ntegrated \textbf{M}ental he\textbf{A}lth da\textbf{T}as\textbf{E}t) dataset that would train DLMs to capture PHQ-9-answerable information from user text. In this research, we restrict our experiments and discussion on whether \textbf{PRIMATE} can help capture context from the user post relevant to some PHQ-9 questions and  pointing out which other PHQ-9 questions would form candidates to direct FQ generation. Our approach and insights have applications to Anxiety (GAD-7), Suicide (C-SSRS), and other mental health disorders as well.



\section{Related Work}
Recently, DLMs have attracted much attention for question answering, thanks to their successes in NLP applications \cite{DBLP:journals/corr/abs-2201-08239, DBLP:journals/corr/abs-2112-04426}. Research on question generation has focused on improving the legibility and relevance of questions. This is because DLMs continue to hallucinate while generating questions in general-purpose domains, which can lead to factually incorrect responses. This can have severe consequences in the mental health domain \cite{DBLP:journals/corr/abs-2201-08239}. Recently, inappropriate and toxic behaviors of language models have been extensively studied and reported in the literature \cite{DBLP:journals/corr/abs-2107-03451, DBLP:journals/corr/abs-2112-04359}. Solutions around fine-tuning, augmenting a neural retriever to support generation, and rules on generation quality have been defined as possible remedies \cite{manas2021knowledge}. These have been effective for the general-purpose domain; however, the research surrounding DLMs is yet to unfold in mental health. ELIZA \cite{10.1145/357980.357991} could transform users' statements into questions but employs labor-intensive templates to generate safe and relevant questions. Models like RAG and REALM were developed to include external knowledge to support question generation \cite{lewis2020retrieval, guu2020realm}. However, these models are still susceptible to incoherent and irrelevant FQ generation . 
Further, their end-to-end learning approach is rigid to support process-guided question generation and discourse, often followed in a clinical setting for triage \cite{gaur2021iseeq}.  

In theory, DLMs should be capable enough of extracting pieces of information from user description that portrays the understanding of the user and leverage it for generating the next FQ. For such a task, supervised training of DLMs with process knowledge and coupling it with information retrieval over domain-specific mental health knowledge is a viable solution. This is because mental health knowledge sources (e.g., SCID (Structured Clinical Interviews for DSM-5) have structured/semi-structured information on how interviews are performed \cite{brodey2018rapid}. Our research substantiates that DLMs (e.g., T5) generate low quality follow-up questions in the context of depression for triage, and granting external knowledge through PHQ-9 reduces the rate at which models generate meaningless FQs \cite{DBLP:journals/corr/abs-2201-08239, DBLP:journals/corr/abs-2107-07566}. In the current research, we define an approach for supervised training of DLMs on a specific dataset that would yield probability distribution over PHQ-9 (with support from Extended PHQ-9). These probabilities will confirm whether the DLM can identify cues from user text that can inform a set of PHQ-9 questions. Remaining PHQ-9 questions are potential FQs.

\paragraph{Datasets:} Prior datasets such as
Counsel Chat \cite{counselchat}, Counseling Conversations \cite{huang2015language}, Role Play \cite{demasi2019towards}, Crisis Text Line \cite{althoff2016large} and Reddit C-SSRS \cite{gaur2019knowledge} have been created to train CA for mental health counseling. Trained CAs can engage in a single turn question answering; however, conducting a conversation requires capturing user context and leveraging clinical instruments to guide the generation of FQs.
\begin{table*}[!ht]
\footnotesize
\begin{center}
\begin{tabular}{p{5.cm}p{5.cm}p{5.cm}}
    \toprule[1.5pt]
      \textbf{Model 1 QGs} & \textbf{Model 2 QGs} & \textbf{Model 3 QGs} \\ \midrule

\tabitem How long ago did the worsening chronic fatigue continue to cripple my life? \textit{[*User already mentions in the title of the post he has been facing the fatigue for years]}
\newline
\tabitem 
What was the estimated dose of suicide? \textit{[*Unsafe incoherent question]}
\newline
\tabitem 
How could body parts feel far less hot in other parts of my life? \textit{[*Incoherent question]}
\newline
\tabitem 
What has caused the gradual cognitive decline? \textit{[*It’s the physician who needs to answer this question after the conversation]}
\newline
\tabitem 
What effect is that I'm less optimistic, and it's embarrassing ? \textit{[*Incoherent attempt to generate a question from the sentence in the post which mentions “I am still pretty optimistic , but I 'm just too out of it , it 's embarrassing”]}
\newline
\tabitem 
What was the reason that increasing depression has a negative effect on me? \textit{[*Irrelevant and incoherent]}
\newline
\tabitem 
A sleep study consists of what? \textit{[*Irrelevant to the diagnostic procedure]}
\newline
\tabitem 
What did Caffeine feel in my body? \textit{[*Wrong pronoun used and irrelevant to the diagnosis]}

   &

\tabitem Have you ever tried many depressants or mental health things? \textit{[*Redundant question as user has already mentioned in the post text that he has tried antidepressants]}
\newline
\tabitem 
Are you seeing a therapist? \textit{[*Redundant question as user had already mentioned in the post text that he has seen a therapist in the past]}
\newline
\tabitem 
Do you even ever try to focus on negative things and/or pain, relief, and patience? \textit{[*Incoherent question]}
\newline
\tabitem 
And what if you forgot to fucking realize that, then it will be fucking so much that you'll be fucking fucking can we ask what they want. \textit{[*Unsafe objectionable question]}
\newline
\tabitem 
Do you mean heavy bipolar disorder? \textit{[*Irrelevant to the context of the post]}
\newline
\tabitem 
Are you aware about the qualities of antidepressants? \textit{[*Inquisitive in nature but irrelevant from a diagnostic perspective]}
\newline
\tabitem 
What does antidepressant work at? \textit{[*Irrelevant from a diagnostic perspective]}
\newline
\tabitem 
Do you have a close friend you can go to who you can talk with, that way you can get out of the house? \textit{[*FQ]}

& 

Are you on any antidepressants? \textit{[*User mentions in the post he has tried them before but no mention has been made if he is still on them]}
\newline
\tabitem 
Do you have any positive thoughts? \textit{[*FQ]}
\newline
\tabitem 
Are you in any danger of hurting yourself? \textit{[*Slightly unsafe]}
\newline
\tabitem 
Is it that you aren't happy with your feelings? \textit{[*Irrelevant question]}
\newline
\tabitem 
Have you tried some exercise? \textit{[*Redundant question as user already mentions he has tried it]}
\newline
\tabitem 
Do you wake up frequently? \textit{[*FQ]}
\newline
\tabitem 
How is your sleep quality? \textit{[*FQ]}
\newline
\tabitem 
When you wake up, what do you do? \textit{[*FQ]}
\newline
\tabitem 
Is there anything that helps you calm the symptoms for now? \textit{[*FQ]}
\newline
\tabitem 
What are your hobbies? \textit{[*Generic FQ]}
\newline
\tabitem 
What are your interests? \textit{[*Generic FQ]}
\\

     \bottomrule[1.5pt]
\end{tabular}
\end{center}
\caption{Examples of questions generated by T5 when tasked to generate FQs when the user query for the \textbf{post} in Figure \ref{fig:motivate} was provided as input. \textbf{Model 1}, which is a pre-trained T5 \cite{raffel2019exploring}, often generates questions which are irrelevant, unsafe, incoherent, and redundant. \textbf{Model 2}, which is T5 fine-tuned on r/depression\_help seems to be relatively coherent and inquisitive compared to \textbf{Model 1}. However, both models generate questions about the topic that user has discussed in their query. As a result, we see that pre-trained and fine-tuned DLMs fail to generate FQs. By enforcing FQ generation using using a dataset curated using extended PHQ-9, generated questions have been mostly inquisitive. This is shown by \textbf{Model 3}. Still, a lot of generations are around the problem the user mentioned.}
 \vspace{-1em}
\label{tab:example1}
\end{table*}



\begin{table*}[!ht]
\footnotesize
\begin{center}
\begin{tabular}{p{4.cm}p{3.cm}p{3.5cm}p{3.5cm}}
    \toprule[1.5pt]
     \textbf{Post text} &  \textbf{Model 1 QGs} & \textbf{Model 2 QGs} & \textbf{Model 3 QGs} \\ \midrule

I feel depressed . I have felt depressed for a while. I have done a lot of things to try and help myself \textit{through therapy} ,[...] , staying busy , going to university . all I have found is that each time I complete any of these things that I still feel like I am empty . When I do not do those things for more than 10 mins by myself that my thoughts just get dark and I be little everything I have ever done . [....] Also I get depressed because I have dreams where I 'm constantly bombarded by women who are not my wife who want to have relationships with me and it 's a nightly battle to not give in to it when I 'm dreaming . I know it sounds silly but I fear that if I give into the dreams that I might do it in real life . \textit{I do n't want to tell my wife} because she does n't understand other weird dreams that I have had in the past .

& 
\tabitem What if I do not do it for more than 10 mins by myself? \textit{[*Closed book question already answered in the post and irrelevant to the diagnostic procedure]}
\newline
\tabitem
What might happen to any women that want to have relationships with me? \textit{[*Irrelevant to the diagnostic process]}
\newline
\tabitem
What do the women at work do?  \textit{[*Irrelevant to the diagnostic process)]}
\newline
\tabitem
What kind of dreams are I was depressed? \textit{[*Closed book question generated directly from a statement in the post]}

&

Would you be happy with them? \textit{[*Irrelevant to the diagnostic process)]}
\newline
\tabitem
What's your job? \textit{[*Irrelevant]}
\newline
\tabitem
Are you able to reach out to your doctor or psychiatrist? \textit{[*User has already mentioned in the original post that he has talked to his therapist]}
\newline
\tabitem
what are your dreams? \textit{[*Closed book question already answered in the post]}
\newline
\tabitem
What do you enjoy about your dreams? \textit{[*FQ]}
\newline
\tabitem
Do you want to talk to people about what dreams are ? \textit{[*FQ]}
\newline
\tabitem
Do you feel as satisfied as yourself when you are dreams? \textit{[*FQ]}

& 

Have you talked to a therapist? \textit{[*Already answered by the user in the post itself]}
\newline
\tabitem
Is there anything that helps you calm your feeling for now? \textit{[*FQ]}
\newline
\tabitem
``Have you ever gotten any help?" \textit{[*Already answered in the query]}
\newline
\tabitem
Do you feel isolated? \textit{[*FQ]}
\newline
\tabitem
What are your hobbies? \textit{[*FQ]}
\newline
\tabitem
What are your interests? \textit{[*FQ]}
\newline
\tabitem
How long have you been waiting for your wife to talk about these dreams? \textit{[*FQ]}
\newline
\tabitem
Have you told your wife you're depressed or not? \textit{[*Inquisitive in nature but already answered by the user in original post]}
\\
     \bottomrule[1.5pt]
\end{tabular}
\end{center}
\caption{In this example, the generated questions from both Model 2 and Model 3 seem to be relevant FQs, but they are not assessing the severity of the mental health condition, despite Model 3 being fine-tuned on a dataset filtered by PHQ-9 questions. In comparison to the qualitative outcome in Table \ref{tab:example1}, this showcases the inability of T5 to support mental health triage.}
\vspace{-1em}
\label{tab:example2}
\end{table*}

\section{Question Generation (QG)}
\paragraph{Dataset for QG:} Our approach to data collection involves scraping posts and comments from r/depression\_help, 
a subreddit on Reddit, which is meant to provide advice and support to help individuals suffering from depression. The posts on this subreddit contain flair tags such as \textit{SEEKING HELP}, \textit{SEEKING ADVICE}, and \textit{REQUESTING SUPPORT}. We filter down the data curated from this subreddit based on the flair tag attribute to retain only \textit{advice, help} or \textit{support} seeking posts and their comments. After filtering, our dataset had approximately 21,000 posts. Each post contains a title, description, and comments. On average, each post has 5 comments. Next, we chunked the main text of each post into smaller groups of sentences (chunks) of less than 512 tokens while making sure no sentence is segmented in between. The motivation for chunking is to ensure no context is lost from the post due to the limitation of T5 to process 512 tokens as input (DLMs in general suffer from such representation limits). We also appended the post title to each chunk to ensure that main idea of each post was captured in it's chunks. This curated dataset tests T5's capability to generate FQs similar to any of the questions in the extended PHQ-9 questionnaire.

\paragraph{Extending PHQ-9 to support FQ generation:}
PHQ-9 questions are subject to different interpretations depending on patient-MHP interaction. Additionally, nine questions are limited in scope for use in tasks like fine-tuning and similarity-based performance evaluations. 
Therefore, to increase the strength of PHQ-9, we collaborated with MHPs to create sub-questions for each question in PHQ-9. First, we used Google SERP API\footnote{\url{https://serpapi.com/}} and Microsoft Bing Search API\footnote{\url{https://www.microsoft.com/en-us/bing/apis/bing-web-search-api}} to retrieve ``People-Also-Ask'' questions. For each question, we retrieved 40 questions by manually searching and assessing their relevance to PHQ-9 questions. Next, we provided the set of 360 questions to three MHPs for assessment. MHPs evaluated the questions on two grounds:(a) Whether they would ask such a question to a patient? (relevance) (b) If yes, when should such a question be asked? (rank). Based on their ratings, we created a final set of 134 sub-questions for the nine questions in PHQ-9\footnote{Questions in extended PHQ-9 : \href{https://docs.google.com/spreadsheets/d/1w6MJoJF1SJ0Bgecw0QUwS\_cqEkUFV8fAp\_D9ULmXUAY/edit\#gid=170478939}{link}} resulting in a total of 143 questions.

\paragraph{Models for FQ Generation:}
We used an off-the-bench T5-base QG model that was fine-tuned on the SQuAD 2.0 question generation dataset \cite{DBLP:journals/corr/abs-1806-03822}~\textbf{[Model 1]}. Next, we fine-tuned Model 1 on r/depression\_help posts and comments. To align with our task of making T5 generate relevant FQs, we filtered out comments which were non-interrogative. We kept only the interrogative statements asked by Reddit users in the comments ~\textbf{[Model 2]}. 
Not all interrogative comments by Reddit users are \textit{diagnostically relevant} FQs (Eg: ``Can you use MS Excel?'', ``Were you interactions on FaceTime?''). To remove such questions, we further filtered the dataset by calculating the maximum BLEURT score between the question (present in the comments) and the questions in extended PHQ-9. We applied a threshold of $0.60$ to this score\footnote{empirically judged}. This removed harmful and diagnostically irrelevant questions while preserving contextual, semantically relevant, and legible questions \textbf{[Model 3]}. See Fig \ref{fig:motivate} for examples of diagnostically relevant questions.

\begin{table*}[!ht]
    \centering
   \begin{tabular}{lccc c ccc}
\toprule[1.5pt]
  $|\hat{Q}|$($\downarrow$) & \multicolumn{3}{c}{Hit Rate on BLEURT} && \multicolumn{3}{c}{Hit Rate on Rouge-L} \\ \cmidrule{2-4} \cmidrule{6-8}
 $\delta$($\rightarrow$) & 0.4 & 0.5 & 0.7 && 0.4 & 0.5 & 0.7 \\ \midrule[1pt]
\multicolumn{8}{c}{\textbf{Model 1}: Pre-trained T5} \\ \midrule[1pt]
5 & 0.5417 & 0.1233 & 0.0020 && 0.1241 & 0.0386 & 0.0005 \\
10 & 0.5400 & 0.1203 & 0.0010 && 0.1290 & 0.0400 & 0.0010 \\ 
15 & 0.5368 & 0.1250 & 0.0013 && 0.1266 & 0.0384 & 0.0009 \\ \midrule[1pt]
\multicolumn{8}{c}{\textbf{Model 2}: Fine-Tuned T5 on r/depression\_help} \\ \midrule[1pt]
5 & 0.6657 & 0.2804 & 0.0097 && 0.3445 & 0.1560 & 0.0100 \\
10 & 0.6691 & 0.2792 & 0.0104 && 0.3481 & 0.1590 & 0.0098 \\
15 & 0.6726 & 0.2787 & 0.0104 && 0.3476 & 0.1588 & 0.0094 \\ \midrule[1pt]
\multicolumn{8}{c}{\textbf{Model 3}: T5 Fine-tuned on r/depression\_help filtered by PHQ-9} \\ \midrule[1pt]
5 & 0.9489 & 0.7088 & 0.1261 && 0.7457 & 0.4937 & 0.0903 \\
10 & \textbf{0.9542} & \textbf{0.7126} & 0.1272 && 0.7460 & \textbf{0.5002} & \textbf{0.0947} \\
15 & 0.9514 & 0.7098 & \textbf{0.1274} && \textbf{0.7484} & 0.4945 & 0.0916 \\
\bottomrule[1.5pt]
\end{tabular}
    \caption{Experimental results comparing different models in generating questions that match the sub-questions in PHQ-9. $\hat{Q}$ is the set of generated questions in each chunk. The performance is recorded over all the generated questions ($\mathbf{\hat{Q}}$). $\delta$ was used as the threshold on the similarity between generated question and PHQ-9 sub-questions while calculating hit rate. BLEURT records semantic similarity, whereas Rouge-L records the longest common subsequence exact match between generated question and PHQ-9 sub-questions. The highest performance on semantic and string similarity is bolded. Acceptable performance in Model 3 achieved using PHQ-9 motivated us to prepare \textbf{PRIMATE}.}
    \label{tab:my_label}
\end{table*}

\begin{figure*}[!ht]
    \centering
    \includegraphics[width=120mm, scale=0.50]{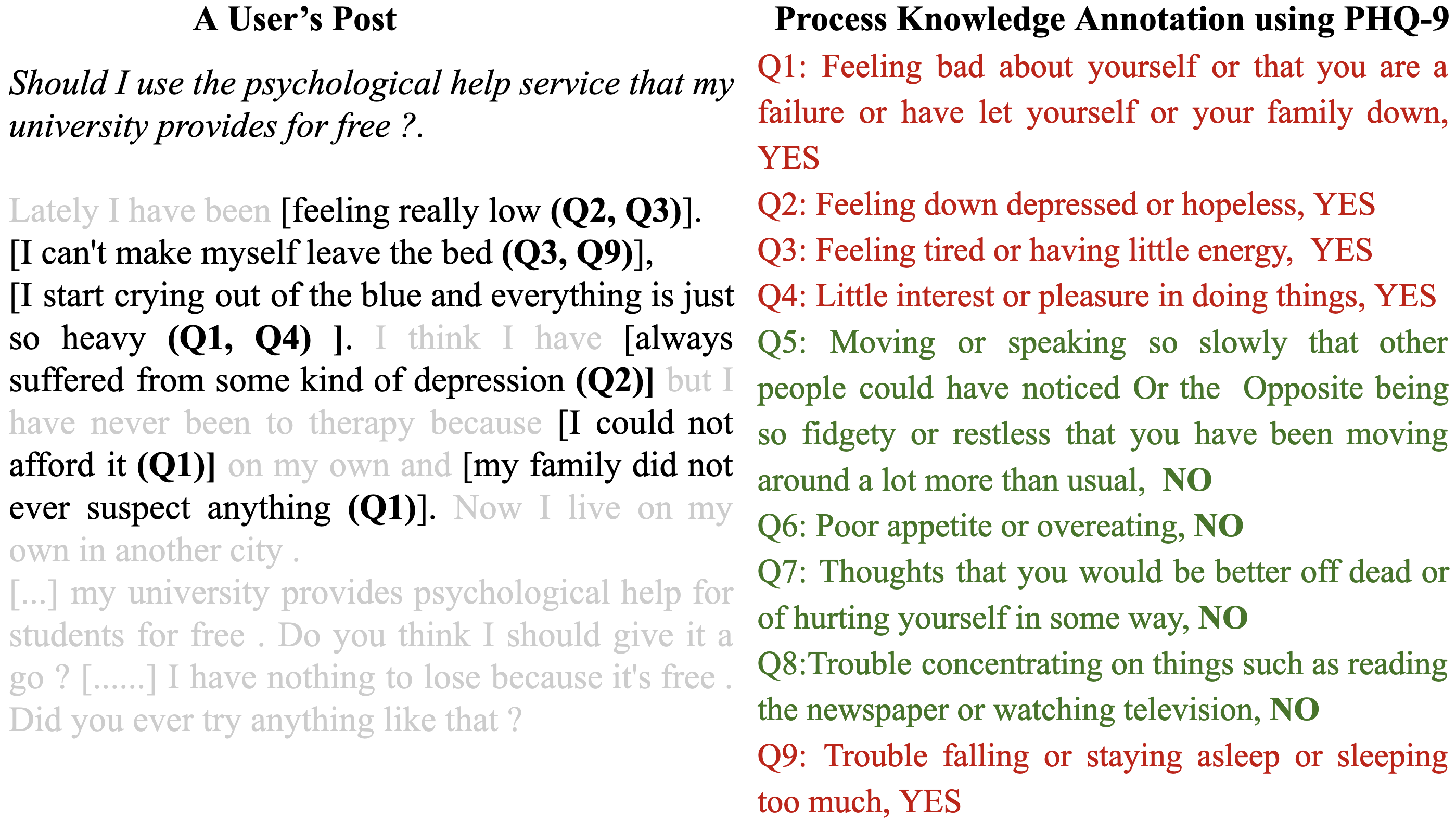}
    \caption{A post in \textbf{PRIMATE} which is annotated with PHQ-9.The questions marked ``YES'' are answerable by DLMs using  the mental health specific cues from user text. The questions marked ``NO'' are the questions a DLM should consider asking as FQs. Sentences within \textbf{[]} were taken as signals that the ``YES'' marked questions had already been answered in the post .}
    \label{fig:primate}
\end{figure*}

\paragraph{Analysis of Models for Question Generation:}
Out of the 21k questions, performance of Models 1, 2, and 3 were examined on those 2003 posts that had at least one interrogative comment. Each of the three models was made to generate FQs in sets of 5, 10, and 15 through nucleus sampling \cite{holtzman2019curious}. For a generated question, BLEURT score was computed with each question in Extended PHQ-9 and the maximum among those scores was taken as the score for the generated question. A clear distinction between models 1, 2, and 3 is the nature of the questions asked. Model 1 generated closed book questions, whereas Model 2 and 3 seem to show some inquisitive nature and seem more focused on the mental health domain, which can be attributed to the after effect of finetuning on Reddit (see Table \ref{tab:example1} and \ref{tab:example2}). We captured the performance of the models quantitatively using 'hit rate' as a metric. For a generated question ($\hat{q}$), we denote :
    \begin{align}
      &score(\hat{q}) = max(bleurt\_score(\hat{q},q_1), \nonumber \\  &bleurt\_score(\hat{q},q_2),...,bleurt\_score(\hat{q},q_{143})) \nonumber,
    \end{align}
where $q_{1}, q_2 \text{...}q_{143} \in\text{Extended-PHQ-9}$. Across all 2003 posts, we had \(C\) = 2575 chunks\footnote{Chunking was done as DLM accepts a maximum input length of 512 tokens.}. Let total number of questions generated by a model be $|\mathbf{\hat{Q}}|$ and $|\hat{Q}|$ denote the number of question generated by the model for a given chunk. For experimentation, we set $|\hat{Q}|$ to have values \{5, 10, 15\}.  Thus, $|\mathbf{\hat{Q}}| = |\hat{Q}| * C$. Then the \textbf{Hit Rate} for a model was computed as: 
\vspace{-1em}
\[
\text{Hit Rate(model}, |\hat{Q}|) = 
\frac{\sum\limits_{\hat{q}~\epsilon~\mathbf{\hat{Q}}} \mathbf{I}(score(\hat{q})>\delta)}{|\mathbf{\hat{Q}}|},
\]
where $\delta$ is the threshold on the similarity between generated question in a chunk and sub-questions in PHQ-9 and $I[\varphi]$ is the indicator function taking values 0 or 1 for a predicate $\varphi$ (Table \ref{tab:my_label} has the scores).

\paragraph{Inference:} 
\textbf{(1)} Regardless of fine-tuning and filtering based on PHQ-9 questions, inherently, T5 does not capture the meaning and usage of the words in the mental health context. Moreover, T5 fails to generate legible and relevant FQs as safe as PHQ-9 questions. Therefore, we scrutinize the generated FQs by mapping them to most similar questions in extended PHQ-9. Examples of irrelevant  generations by T5 that it thought were relevant are: (a) ``Wtf?'' (generated FQ) was found most similar to ``Do you have hope?'' (PHQ-9) (b) ``What did Boyfriend suffocate me with during his break up a week after I got a diagnosis?'' (generated FQ) was found most similar to ``What do you think makes you a failure'' (PHQ-9). The previous generated question is redundant as the answer to it was already present in the original post. \textbf{(2)} Many generated questions contain extreme language due to the informal nature of the Reddit platform, which is very sensitive issue, especially in the mental health domain. Examples are: ``Did you f***ing realize that f***ing people are f***ing too?'' (generated FQ) was found to be the most similar to ``What do you think makes you a failure?''. Thus, T5 and its variants need to capture ``what the user knows and has already mentioned in his post'' by checking which PHQ-9 questions are already answerable using the user's post before generating the next probable FQs in order to avoid redundancy.

\begin{figure*}[t]
    \centering
    \includegraphics[width=\linewidth, scale=0.5]{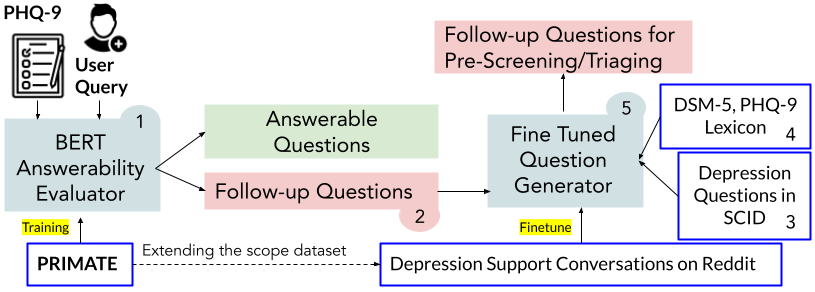}
    \caption{1. Answerability evaluator: A BERT model is trained in a supervised setting to be an evaluator of whether a PHQ-9 question can be answered in a given user post (binary) using PRIMATE. For nine PHQ-9 questions, we require nine such evaluators. 2. Follow up questions: PHQ-9 questions that are not already answerable using the user post form candidates for follow up. 3. SCID: Corresponding to each PHQ-9 question, the SCID describes a clinician approved sub-sequence of questions to obtain the answer to the follow up question. 4. Use existing PHQ-9 and DSM-5 lexicons \cite{yazdavar2017semi} to filter the question to be generated. 5. Generate FQs using T5 fine-tuned on external domain-specific knowledge and the large-scale depression support conversation dataset created from Reddit and \textbf{PRIMATE}.}
    \vspace{-1em}
    \label{fig:proposed_method}
\end{figure*}

\section{PRIMATE for FQ Generation}
We conceptualize our approach on the duality of data and the process knowledge contained in PHQ-9 (see Figure \ref{fig:proposed_method}). First, a BERT Answerability Evaluator identifies which questions in PHQ-9 are already answerable (using the user's initial description of his/her condition in the post) and which ones need more information to be answerable. The latter type of questions form candidates for training a generative DLM for FQ generation. We present \textbf{PRIMATE}, a dataset consisting of Reddit posts containing user situations describing their health conditions and whether the questions in PHQ-9 are answerable using the content in the posts. Each question is attributed with a binary ``yes" or ``no" label stating whether the user's description already contains the answer to that question (see Table \ref{tab:primate_stats}). \textbf{PRIMATE} was created from a month long annotation-evaluation cycle between MHPs and crowd workers. A total of five crowd workers performed this task, achieving an initial annotator agreement of 67\% using Fleiss kappa. Subsequently, the MHPs assessed the quality of annotations and provided their suggestion for improvement, leading to an acceptable agreement score of 85\%. A sample annotated post in \textbf{PRIMATE} is shown in Figure \ref{fig:primate}.

\begin{table}[!ht]
\begin{center}
\begin{tabular}{p{1.5cm}p{2.25cm}p{2.25cm}}
    \toprule
     \textbf{PHQ-9} & \multicolumn{2}{c}{\textbf{Number of Posts}} \\ \cmidrule{2-3} 
     \textbf{Questions} &  \textbf{With Answer (Yes)} & \textbf{W/o Answer (No)} \\ \midrule 
     Q1 & 1679 & 324 \\ \midrule 
     Q2 & 1664 & 339 \\ \midrule 
     Q3 & 686 & 1317 \\ \midrule 
     Q4 & 949 & 1054 \\ \midrule 
     Q5 & 530 & 1473\\\midrule 
     Q6 & 195 & 1808 \\\midrule 
     Q7 & 741 & 1262  \\ \midrule 
     Q8 & 196 & 1807\\ \midrule 
     Q9 & 374 & 1629 \\ \bottomrule
\end{tabular}
\end{center}
\caption{Distribution of 2003 posts in \textbf{PRIMATE} according to whether the text in the post answers a particular PHQ-9 question. Through this imbalance, \textbf{PRIMATE} presents its importance in training DLM(s) to identify potential FQs in PHQ-9 that would guide a generative DLM(s) to conduct a discourse with a patient with a vision to assist MHPs in triage. Q1-Q9 are described in Figure \ref{fig:primate}}
\label{tab:primate_stats}
\end{table}

\paragraph{BERT as Answerability Evaluator:}
While Model 3 shows respectable performance (Table \ref{tab:my_label}), even the FQs generated by Model 3 may not yield the most efficient capture of the PHQ-9 related questions (evident from the low hit rate at a higher threshold) ($\delta$). The MHPs would probably have a more streamlined, focused questioning strategy. For efficient MHPs and AI collaboration, we propose to guide the questioning in a more systematic way by predicting if the user post already has answers to the PHQ-9 questions. This is first posed as a binary classification problem over nine PHQ-9 questions. Thereafter, the approach is to generate questions similar to the PHQ-9 questions that do not have answers in the post. Thus, we train BERT\footnote{BERT end-to-end training perform well compared to baselines Electra\cite{clark2019electra}, and MedBERT\cite{gu2021domain}} (a transformer-based DLM) as a classifier on the \textbf{PRIMATE} dataset. We plan to further use the classification outcome from the BERT model to drive the direction of further questioning with the patient in a more controlled manner. This process can lead to high efficiency and completion of the mental health triaging in as few questions as possible.

\begin{table}[!ht]
    \centering
  \begin{tabular}{lcccc}
\toprule[1.5pt]
  $\delta$ ($\rightarrow$) & \multicolumn{1}{c}{0.5} & \multicolumn{1}{c}{0.7} & \multicolumn{1}{c}{0.9} & Class- \\ \cmidrule{2-4}
 PHQ-9($\downarrow$) & MCC  & MCC  & MCC & Type \\ \midrule[1pt]
\rowcolor{LightCyan}
 Q1 & 0.0 & 0.17 & 0.17 & W  \\
 Q2 & 0.43 & 0.45 & 0.52 & S \\
 Q3 & 0.41 & 0.46 & 0.33 & \textbf{M} \\
 \rowcolor{LightCyan}
 Q4 & 0.14 & 0.19 & 0.13 & W \\
 Q5 & 0.63 & 0.65 & 0.66 & S \\
 \rowcolor{LightCyan}
 Q6 & 0.47 & 0.43 & 0.27 & W \\
 Q7 & 0.66 & 0.68 & 0.7 & S \\
 \rowcolor{LightCyan}
 Q8 & 0.1 & 0.0 & 0.0 & W \\
 Q9 & 0.62 & 0.56 & 0.39 & \textbf{M}\\
\bottomrule[1.5pt]
\end{tabular}
    \caption{We record the Matthews Correlation Coefficient (MCC) to measure the performance of the Evaluator (see Figure \ref{fig:proposed_method}). The MCC score for all 9 questions across different thresholds is in the range 0 to +1 (low to high positive relationships). The MCC for some configurations runs into a divide by zero error, and we replace this value with 0.0. 
     \textbf{W}: model is unable to learn cues to determine answerability in a post.
     \textbf{M}: model is uncertain whether a particular PHQ-9 question is answerable or not.
     \textbf{S}: answerability can be determined by the model with high reliability. Class-Type: Classification Type when $\delta =\:0.9$}
    \label{tab:my_label2}
\end{table}

\paragraph{Performance Analysis:}
We report the Matthews Correlation Coefficient (MCC) scores in table \ref{tab:my_label2}. MCC is a reliable metric to assess a model's classification over an imbalanced dataset, particularly useful when we are interested in all four categories of confusion matrix: true positives (answerable questions (AQ)), true negatives (FQ candidates), and false alarms (false negatives and positives).
As \textbf{PRIMATE} shows a disproportional distribution of AQs (yes) and FQs (no), MCC is an appropriate metric \cite{chicco2020advantages}. We base our analysis on the consistency of BERT classifier on varying threshold ($\delta$) in table \ref{tab:my_label2}. 
A score between 0.0 to 0.30 (Type \textbf{W: Weak}) on MCC means the model is only able to find a negligible to weak positive relationship between input and output. In our context, a score in this range for a particular PHQ-9 question means that model is unable to effectively learn the cues needed to judge the answerability of that question in user posts. 
A score between 0.30 and 0.40 (Type \textbf{M: Maybe}) means that the model is able to learn a moderately positive relationship, interpreted as ambiguity in the model to judge whether a particular PHQ-9 question is answerable from user posts. 
MCC scores between 0.40 to 0.70 (Type \textbf{S: Strong}) for a question in PHQ-9 means that the model can effectively judge whether that question is answerable in user posts . Any score above 0.70 makes the model's judgements even more reliable. This experiment completes steps 1 and 2 in Figure \ref{fig:proposed_method}. Steps 3, 4 and 5 are concerned with the task of FQ generation by fine-tuning the T5 DLM as a generator over r/depression\_help and other depression support communities on Reddit. The FQ generations will be controlled using the process knowledge in SCID which is consulted for interviewing by MHPs. Further, PHQ-9 lexicons are leveraged for promoting diversity and filtering irrelevant FQ generations. 
We leave this process of FQ generations to shape discourse as future work. 

\section{Conclusion}
This paper demonstrated the importance of data and process knowledge to adapt DLMs for generating FQs that would assist MHPs in triaging depression. 
Our experiments show that without process knowledge, DLMs hallucinate by generating unsafe, incoherent, and  irrelevant questions that are not helpful for MHPs in pre-screening or triaging. The challenge lies in the inability of the DLMs to judge from the set of generated questions, which is a potential effective FQ to ask based on the user information. 
The improved question generation performance of DLMs fine-tuned on conversational data filtered by process knowledge encouraged us to prepare PRIMATE. \textbf{PRIMATE} can train DLMs to judge `whether a user’s description of their mental health condition already contains an answer to a particular question in PHQ-9', which would eventually guide coherent FQ generations. We leave our approach for FQ generation as future work, but provide sufficient details on the broader forms of knowledge needed in realizing such a pipeline. 

\paragraph{Limitations:} 
We are yet to scale our understanding to other mental health disorders, such as anxiety using GAD-7 and Suicidality using C-SSRS \cite{jiang2020detection}. Further, we are yet to investigate whether \textbf{PRIMATE}, along with the knowledge in SCID can make DLMs transferable across multiple mental health disorders, especially the ones comorbid with depression. Also, there is a need for a clinically explainable safety metric for our task.

\paragraph{Ethical Considerations:}
Mental health communities on Reddit offer a crowd perspective on various disorders wherein the FQs in the comments highlight the good intentions of Reddit users to help users with conditions, such as depression. We take such interactions as a proxy for improving patient-MHP interactions. \cite{benton2017ethical} described that studies involving user-generated content are exempted from the IRB requirement as long as the data source is public and the user's identity is not recognizable. Apart from being publicly available, Reddit users are anonymous, and we further work with random user IDs. Since we make \textbf{PRIMATE} public for research use, we use a Data Use Agreement \cite{losada-crestani2016} for responsible dissemination of the dataset.

 
 %
 
\section{Acknowledgment}
We acknowledge partial support from the National Science Foundation (NSF) award \#2133842 “EAGER: Advancing Neuro-symbolic AI with Deep Knowledge-infused Learning,” with PI Dr. Amit Sheth.  Any opinions, findings, and conclusions or recommendations expressed in this material are those of the author(s) and do not necessarily reflect the views of the NSF.

\newpage
\bibliographystyle{acl_natbib}
\bibliography{acl2021.bib}


\end{document}